\documentclass[runningheads]{llncs}

\newif\ifdebug

\usepackage{cite} 
\usepackage{booktabs}
\usepackage{blindtext}
\usepackage[linesnumbered,lined,ruled,vlined]{algorithm2e}
\SetAlFnt{\scriptsize}

\SetCommentSty{mycommfont}
\usepackage{capt-of}
\usepackage{wrapfig} 
\usepackage[caption=false]{subfig} 
\usepackage{epstopdf} 
\usepackage{floatrow} 
\usepackage[percent]{overpic}
\usepackage[table]{xcolor}
\usepackage{tabularx}
\usepackage{setspace}
\usepackage{float}
\usepackage{textcomp}
\usepackage{epsfig}
\usepackage{esvect}
\usepackage{graphicx} 
\graphicspath {{Figures/}}
\usepackage{tikz} 
\usepackage{comment} 
\usepackage{amsmath,amssymb} 
\usepackage{xfrac} 
\usepackage{color} 
\usepackage{bibentry}
\usepackage[inline]{enumitem} 
\usepackage[normalem]{ulem}
\usepackage[symbol]{footmisc}

\nonstopmode

\usepackage[pdfdisplaydoctitle, bookmarksdepth=2]{hyperref}

\def\argmin{\operatornamewithlimits{\rm arg\,min}}

\makeatletter\@ifundefined{etal}{\def\etal{\textit{et~al.}}}{}\makeatother

\hypersetup{
pdftitle={Efficiency in Real-time Webcam Gaze Tracking},
pdfauthor={Amogh Gudi, Xin Li, Jan van Gemert}
}

\begin{document}
\pagestyle{headings}
\mainmatter
\def\ECCVSubNumber{1}  
\title{Efficiency in Real-time Webcam $\protect\vv{\text{Gaze}}$ Tracking}

\titlerunning{Efficiency in Real-time Webcam Gaze Tracking}
%
\author{Amogh Gudi\inst{1,2}\and 
Xin Li\inst{1,2}\and
Jan van Gemert\inst{2}}
\authorrunning{A. Gudi~\etal}
%
\institute{Vicarious Perception Technologies [VicarVision], Amsterdam, The Netherlands\and
Delft University of Technology [TU Delft], Delft, The Netherlands
\email{\{amogh,xin\}@vicarvision.nl, j.c.vangemert@tudelft.nl}}

\maketitle
	\begin{abstract}
\label{sec:abstract}
Efficiency and ease of use are essential for practical applications of camera based eye/gaze-tracking. 
Gaze tracking involves estimating where a person is looking on a screen based on face images from a computer-facing camera. 
In this paper we investigate two complementary forms of efficiency in gaze tracking: 
1. The computational efficiency of the system which is dominated by the inference speed of a CNN predicting gaze-vectors; 
2. The usability efficiency which is determined by the tediousness of the mandatory calibration of the gaze-vector to a computer screen.  
To do so, we evaluate the computational speed/accuracy trade-off for the CNN and the calibration effort/accuracy trade-off for screen calibration. 
For the CNN, we evaluate the full face, two-eyes, and single eye input. 
For screen calibration, we measure the number of calibration points needed and evaluate three types of calibration: 1. pure geometry, 2. pure machine learning, and 3. hybrid geometric regression. 
Results suggest that a single eye input and geometric regression calibration achieve the best trade-off.
\end{abstract}

	\section{Introduction}
\label{sec:intro}
\begin{figure*}
	\centering
	\includegraphics[width=\textwidth,trim={0 5cm 4.2cm -2.8cm},clip]{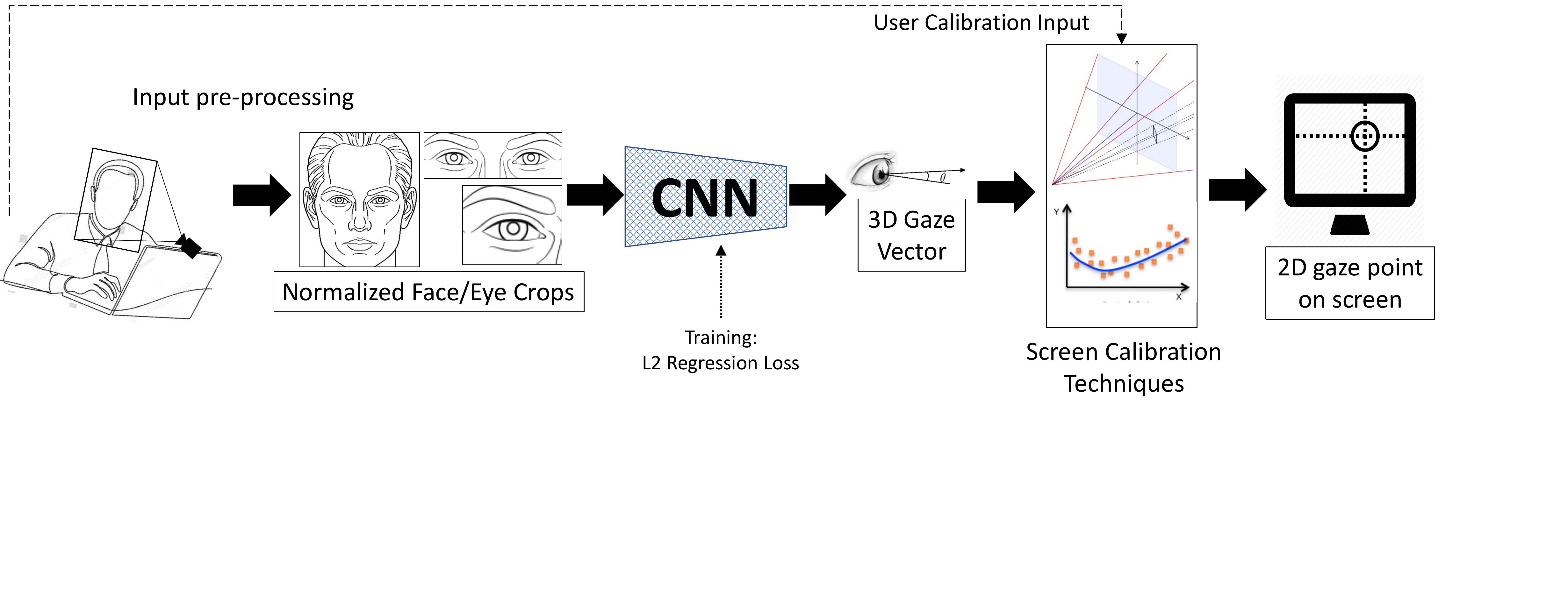}
    \caption{An overview illustration of our camera-to-screen gaze tracking pipeline under study. 
    (Left to right) Images captured by the webcam are first pre-processed to create a normalized image of face and eyes. 
    These images are used for training a convolutional neural network (using L2 loss) to predict the 3D gaze vector. 
    With the cooperation of the user, the predicted gaze vectors can finally be projected on to the screen where he/she is looking using proposed screen calibration techniques. 
    }
	\label{fig:overview}
\end{figure*}

In a typical computer-facing scenario, the task of gaze-tracking involves estimating where a subject's gaze is pointing based on images of the subject captured via the webcam.
This is commonly in the form of a gaze vector, which determines the pitch and yaw of the gaze with respect to the camera~\cite{MPII}.
A more complete form of gaze tracking further extends this by also computing at which specific point the subject is looking at on a screen in front of the subject~\cite{gazecapture,written-on-your-face}.
This is achieved by estimating the position of the said screen w.r.t. the camera (a.k.a. screen calibration), which is not precisely known beforehand.
We present a study of some core choices in the design of gaze estimation methods in combination with screen calibration techniques (see Figure~\ref{fig:overview}), leaning towards an efficient real-time camera-to-screen gaze-tracking system.

\subsubsection*{Computational efficiency: Input size.}
Deep networks, and CNNs in particular, improved accuracy in gaze estimation where CNN inference speed is to a large extent determined by the input image size. The input size for gaze estimation can vary beyond just the image of the eye(s)~\cite{Pictorial,MPII}, but also include the whole eye region~\cite{written-on-your-face}, the whole face, and even the full camera image~\cite{gazecapture}. Yet, the larger the input image, the slower the inference speed. We study the impact of various input types and sizes with varying amounts of facial contextual information to determine their speed/accuracy trade-off.

\subsubsection*{Usability efficiency: Manual effort in screen calibration.}
Most work focus on gaze vectors estimation~\cite{RT-GENE, dilated_conv_gaze, Recurrent_Gaze, park2019few, yu2019improving}. However, predicting the \emph{gaze-point}, point on a screen in front of the subject where he/she is looking, is a more intuitive and directly useful result for gaze tracking based applications, especially in a computer-facing/human-computer interaction scenario.  
If the relative locations and pose of the camera w.r.t to the screen were exactly known, projecting the gaze-vector to a point on screen would be straightforward.
However, this transformation is typically not known in real-world scenarios, and hence must also be implicitly or explicitly estimated through an additional calibration step. 
This calibration step needs to be repeated for every setup. 
Unlike gaze-vector prediction, you cannot have a ``pre-trained" screen calibration method. 
In practice, every-time a new eye-tracking session starts, the first step would be to ask the user to look at and annotate predefined points for calibration. Therefore, obtaining calibration data is a major usability bottleneck since it requires cooperation of the user every time, which in practice varies. 
Here, we study usability efficiency as a trade-off between the number of calibration points and accuracy.

We consider three types of calibration. Geometry based modelling methods have the advantage that maximum expert/geometrical prior knowledge can be embedded into the system. 
On the other hand, such mathematical models are rigid and based on strong assumptions, which may not always hold.
In contrast, calibration methods based on machine learning require no prior domain knowledge and limited hand-crafted modelling. However, they may be more data-dependent in order to learn the underlying geometry.
In this paper, we evaluate the efficiency trade-off of various calibration techniques including a hybrid approach between machine learning regression and geometric modelling. 

\subsubsection*{Contributions.}
We have the following three contributions:
\begin{enumerate}[label=(\roman*)] 
    \item We evaluate computational system efficiency by studying the balance of gains from context-rich inputs vs their drawbacks.
    We study their individual impact on the system's accuracy w.r.t. their computational load to determine their efficiency and help practitioners find the right trade-off.
    \item We demonstrate three practical screen calibration techniques that can be use to convert the predicted gaze-vectors to points-on-screen, thereby performing the task of complete camera-to-screen gaze tracking.
    \item We evaluate the usability efficiency of these calibration methods to determine how well they utilize expensive user-assisted calibration data.
    This topic has received little attention in literature, and we present one of the first reports on explicit webcam-based screen calibration methods.
\end{enumerate}

	\section{Related Work}
\label{sec:relwork}
Existing methods for gaze tracking can be roughly categorized into model-based and appearance-based methods.
The former~\cite{EyeTab,3dmm} generates a geometric model for eye to predict gaze, while the latter~\cite{tan2002appearance} makes a direct mapping between the raw pixels and the gaze angles. 
Appearance driven methods have generally surpassed classical model-based methods for gaze estimation.

\subsubsection*{Appearance-based CNN gaze-tracking.}
As deep learning methods have shown their potentials in many areas, some appearance-based CNN networks are shown to work effectively for the task of gaze prediction.

Zhang \etal~\cite{in_the_wild,MPII} proposed the first deep learning model for appearance-based gaze prediction. Park \etal~\cite{Pictorial} proposed a combined hourglass~\cite{hourglass} and DenseNet~\cite{DenseNet} network to take advantage of auxiliary supervision based on the gaze-map, which is two 2D projection binary mask of the iris and eyeball. 
Cheng \etal~\cite{asymmetric} introduced ARE-Net, which is divided into two smaller modules: one is to find directions from each eye individually, and the other is to estimate the reliability of each eye.
Deng and Zhu \etal~\cite{Monocular} define two CNNs to generate head and gaze angles respectively, which are aggregated by a geometrically constrained transform layer.
Ranjan \etal~\cite{Head_Pose_Invariant_Gaze_Tracking} clustered the head pose into different groups and used a branching structure for different groups.
Chen \etal~\cite{dilated_conv_gaze} proposed Dilated-Nets to extract high level features by adding dilated convolution.
We build upon these foundations where we evaluate the speed vs accuracy trade-off in a real-time setting. 
The image input size has a huge effect on processing speed, and we control the input image size by varying eye/face context.

The seminal work of Zhang \etal~\cite{in_the_wild,MPII} utilized minimal context by only using the grayscale eye image and head pose as input.
Krafka \etal~\cite{gazecapture} presented a more context-dependent multi-model CNN to extract information from two single eye images, face image and face grid (a binary mask of the face area in an image).
To investigate how the different face region contributes to the gaze prediction, a full-face appearance-based CNN with spatial weights was proposed~\cite{written-on-your-face}.
Here, we investigate the contribution of context in the real-time setting by explicitly focusing on the speed/accuracy trade-off. 

A GPU based real-time gaze tracking method was presented in~\cite{RT-GENE}. 
This was implemented in a model ensemble fashion, taking two eye patches and head pose vector as input, and achieved good performance on several datasets~\cite{RT-GENE,Image_Normalized,MPII} for person-independent gaze estimation.
In addition, \cite{dilated_conv_gaze,Recurrent_Gaze} have included some results about the improvements that can be obtained from different inputs.
In our work, we perform an ablation study and add the dimension of computation load of each input type. 
Our insight in the cost vs benefit trade-off may help design efficient gaze tracking software that can run real-time beyond expensive GPUs, on regular CPUs which have wider potential in real world applications.

\subsubsection*{Screen calibration: Estimating point-of-gaze.}
In a classical geometry-based model, projecting any gaze-vector to a point on a screen requires a fully-calibrated system. 
This includes knowing the screen position and pose in the camera coordinate system. 
Using a mirror-based calibration technique~\cite{Rodrigues2010CameraPE}, the corresponding position of camera and screen can be attained.
This method needs to be re-applied for different computer and camera setting, which is non-trivial and time-consuming.
During human-computer interactions, information like mouse clicks may also provide useful information for screen calibration \cite{webgazer}.
This is, however, strongly based on the assumption that people are always looking at the mouse cursor during the click.

Several machine learning models are free of rigid geometric modelling while showing good performance.
Methods like second order polynomial regression~\cite{Kasprowski2014} and Gaussian process regression~\cite{GPR} have been applied to predict gaze more universally.
WebGazer~\cite{webgazer} trains regression models to map pupil positions and eye features to 2D screen locations directly without any explicit 3D geometry.
As deep learning features have shown robustness in different areas, other inputs can be mixed with CNN-based features for implicit calibration, as done in \cite{gazecapture, written-on-your-face}. 
CNN features from the eyes and face are used as inputs to a support vector regressor to directly estimate gaze-point coordinates without an explicit calibration step.
These methods take advantage of being free of rigid modelling and show good performance.
On the other hand, training directly on CNN features makes this calibration technique non-modular since it is designed specific to a particular gaze-prediction CNN. 
In our work, we evaluate data-efficiency for modular screen calibration techniques that convert gaze-vectors to gaze-points based on geometric modelling, machine learning, and a mix of geometry and regression.
We explicitly focus on real world efficiency which for calibration is not determined by processing speed, but measured in how many annotations are required to obtain reasonable accuracy.

	\section{Setup}
\label{sec:meth}

The pipeline contains three parts, as illustrated in Figure~\ref{fig:overview}:
\begin{enumerate}
	\item Input pre-processing by finding and normalizing the facial images;
	\item A CNN that takes these facial images as input to predict the gaze vector;
	\item Screen calibration and converting gaze-vectors to points on the screen.
\end{enumerate}

\subsection{Input Pre-processing}
The input to the system is obtained from facial images of subjects.
Through a face finding and facial landmark detection algorithm~\cite{CCNFs}, the face and its key parts are localized in the image. 
Following the procedure described by Sugano~\etal~\cite{Image_Normalized}, the detected 2D landmarks are fitted onto a 3D model of the face.
This way, the facial landmarks are roughly localized in the 3D camera coordinate space.
By comparing the 3D face model and 2D landmarks, the head rotation matrix $\mathbf{R}$ and translation vector $\mathbf{T}$, and the 3D eye locations $\mathbf{e}$ are obtained in 3D camera coordinate space.
A standardized view of the face is now obtained by defining a fixed distance $d$ between the eye centres and the camera centre and using a scale matrix $\mathbf{S} = \text{diag}(1,1,\frac{d}{||\mathbf{e}||})$.
The obtained conversion matrix $\mathbf{M}=\mathbf{S} \cdot \mathbf{R}$ is used to apply perspective warping to obtain a normalized image without roll (in-plane rotation). For training, the corresponding ground truth vector $\mathbf{g}$ is similarly transformed: $\mathbf{M} \cdot \mathbf{g}$.

\subsection{CNN Prediction of Gaze Vectors}
We use a VGG16~\cite{vgg} network architecture with BatchNorm~\cite{DBLP:journals/corr/IoffeS15} to predict the pitch and yaw angles of the gaze vector with respect to the camera from the normalized pre-processed images.

\paragraph{Training.}
Following the prior work in\cite{MPII}, the network was pre-trained on ImageNet~\cite{ILSVRC15}. For all the experiments conducted in this work, we set the following hyperparameters for the training of the network for gaze-vector prediction:
\begin{enumerate}[label=(\roman*)]
	\item Adam optimizer with default settings \cite{kingma2014adam};
	\item a validation error based stopping criteria with a patience of 5~epochs;
	\item learning rate of $10^{-5}$, decaying by 0.1 if validation error plateaus;
	\item simple data augmentation with mirroring and gaussian noise ($\sigma=0.01$).
\end{enumerate}

\paragraph{Inference.}
This trained deep neural network can now make prediction of the gaze vector.
The predicted gaze-vector (in the form of pitch and yaw angles) are with respect to the `virtual' camera corresponding to the normalized images.
The predicted virtual gaze vectors can be transformed back to the actual gaze vector with respect to the real camera using the transformation parameters obtained during image pre-processing.
These vectors can then be projected onto a point on the screen after screen calibration.

\subsection{Screen Calibration: Gaze Vectors to Gaze Points}
To project the predicted 3D gaze vectors (in the camera coordinate space) to 2D gaze-points on a screen, the position of the screen with respect to the camera must be known which is difficult to obtain in real world settings.
The aim of screen calibration is to estimate this geometric relation between the camera and the screen coordinate systems such that the predicted gaze vectors in camera coordinates are calibrated to gaze-points in screen coordinates.
Because we focus on the task of eye-tracking in a computer-facing scenario, we can simplify the setup by making some assumptions based on typical webcam-monitor placement (such as for built-in laptop webcams or external webcams mounted on monitors):
\begin{enumerate}[label=(\roman*)]
	\item the roll and yaw angles between the camera and the screen are 0\textdegree, 
	\item the intrinsic camera matrix parameters are known, and
	\item the 3D location of the eye is roughly known w.r.t the camera (estimated by the eye landmarks in the face modelling step in camera coordinate space). 
\end{enumerate}
With these assumptions in place, we can design user-aided calibration techniques where the user cooperates by looking at predefined positions on the screen.

\section{Screen Calibration Methods}
As calibration is tedious and needs to be performed multiple times, we evaluate efficiency in terms of how much manual effort is required for three calibration versions:
\begin{enumerate}
	\item calibration by geometry;
	\item calibration by machine learning;
	\item calibration by a hybrid: geometry and regression.
\end{enumerate}

\subsection{Geometry-based Calibration}
\begin{figure}
	\centering
	\subfloat[][$y-z$ plane with pitch angles]{\includegraphics[width=.43\textwidth]{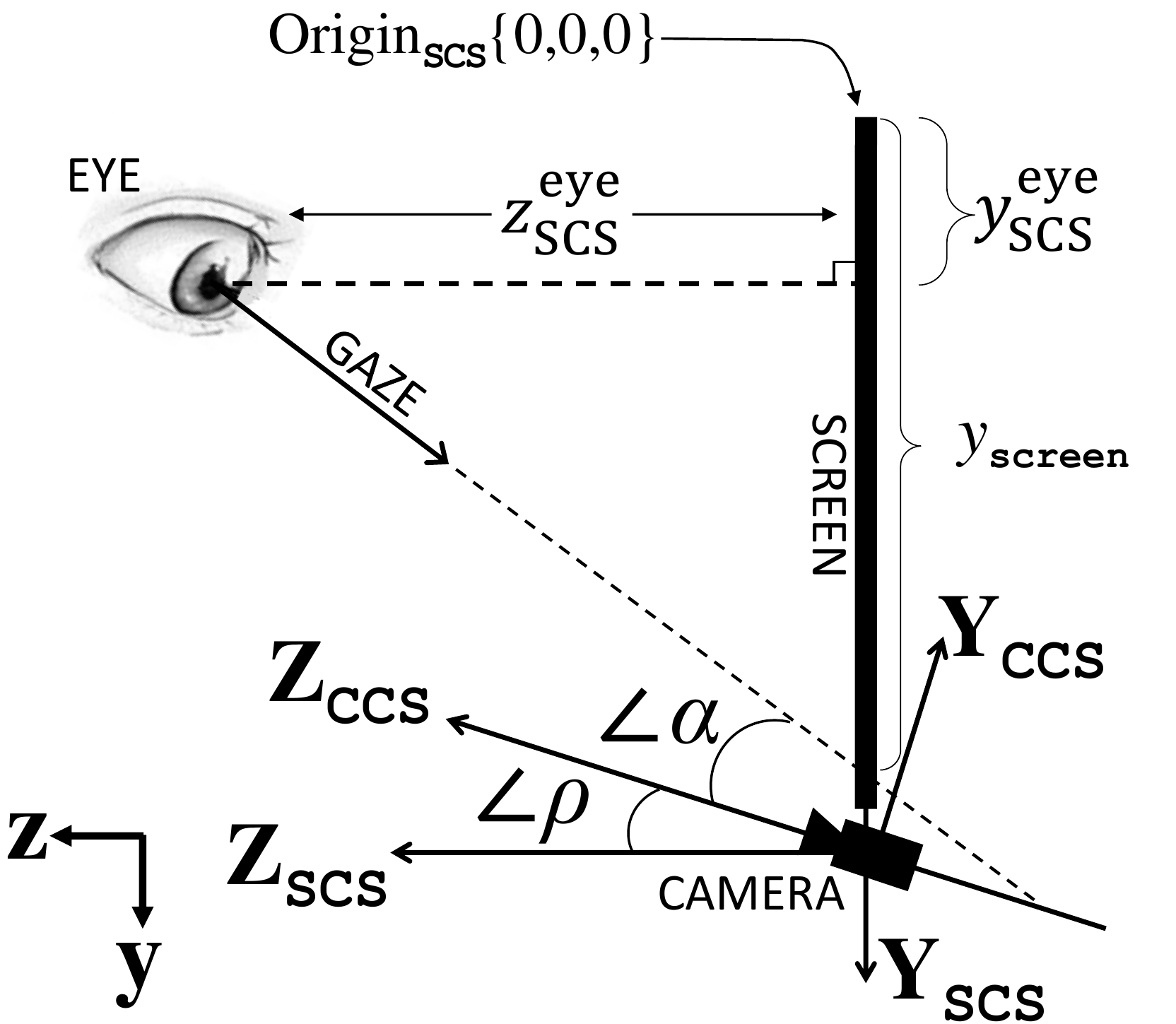}\label{fig:caldrawing-yz}}\qquad
	\subfloat[][$x-z$ plane with yaw angle]{\includegraphics[width=.43\textwidth]{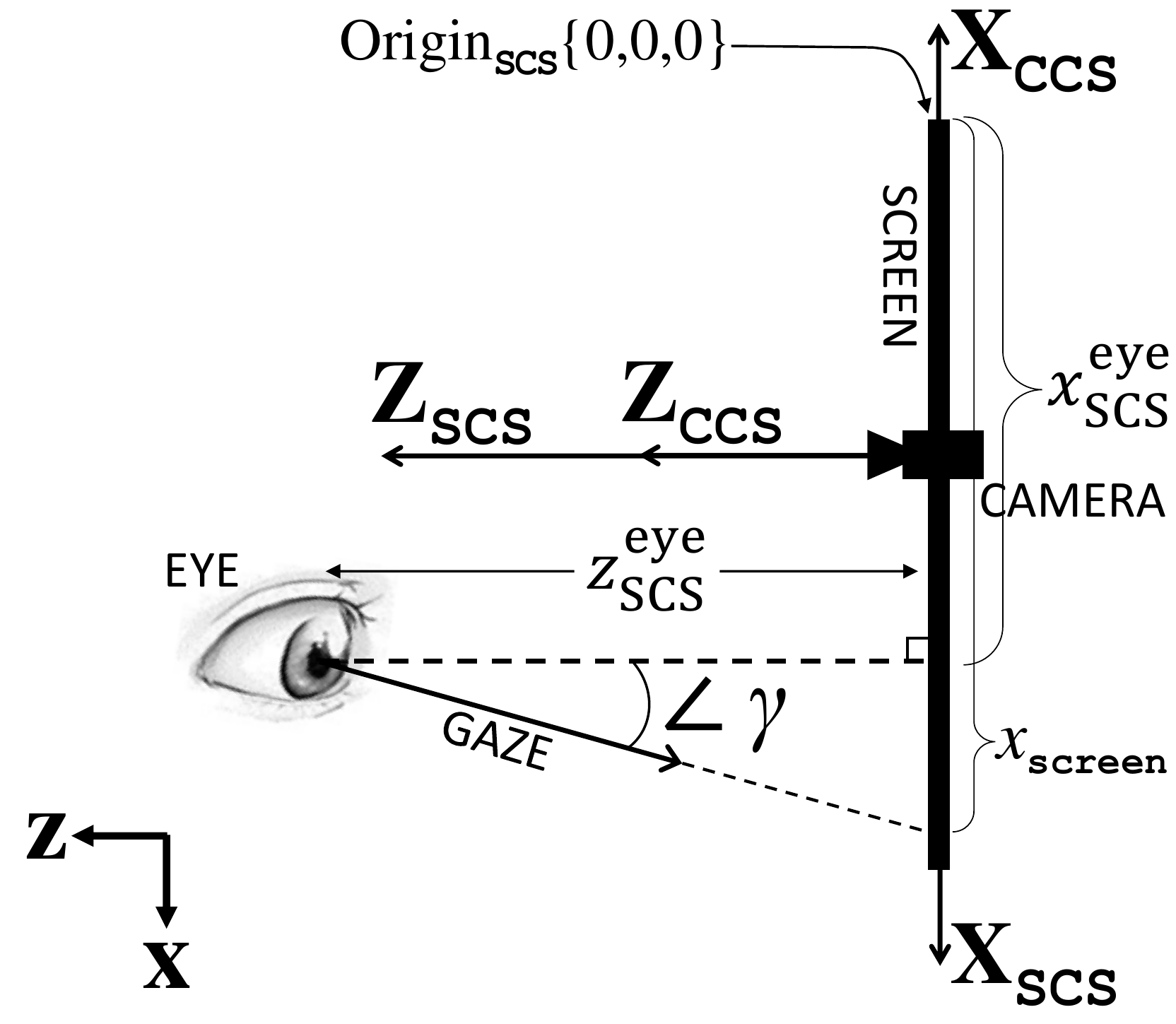}\label{fig:caldrawing-xz}}
	{\caption{A illustration of the geometric setup between the eye and the screen in the screen coordinate space. 
	$\{\mathbf{X},\mathbf{Y}, \mathbf{Z}\}_{\texttt{CCS}}$ and $\{\mathbf{X},\mathbf{Y}, \mathbf{Z}\}_{\texttt{SCS}}$ represent the directions of the $\{x,y,z\}$-axes of the camera and screen coordinate systems respectively; $\text{Origin}_\texttt{SCS}$ represents the origin of the screen coordinate system.
	} 
	\label{fig:caldrawings}}
\end{figure}
To perfectly project a gaze-vector w.r.t the camera to a point on a screen, we are essentially required to determine the transformation parameters between the camera coordinate system (CCS) and the screen coordinate system (SCS). 
With our assumptions about roll and yaw in place, this transformation can be expressed by the rotation matrix $\mathbf{R}$ and the translation vector $\mathbf{T}$ between the camera and the screen:
\begin{equation}
	\mathbf{R} = 	\left[\begin{matrix}
		1 & 0 & 0 \\
		0 & \enskip\cos(\rho) & \enskip-\sin(\rho) \\ 
		0 & \enskip\sin(\rho) & \enskip\cos(\rho) 
	   \end{matrix}\right]
		\quad\&\quad \mathbf{T} = \left[\begin{matrix}
		\Delta x & \Delta y & \Delta z \end{matrix}\right]^T,
	\label{eqn:R}
	\end{equation}
where $\rho$ denotes the vertical pitch angle (about the $x$-axis; along the $y$-axis) between the camera and the screen norm, and $\Delta x, \Delta y, \Delta z$ represent the translational displacement between the camera and the screen. 
An illustration of the geometric setup is shown in Figure \ref{fig:caldrawings} (\ref{fig:caldrawing-yz} and \ref{fig:caldrawing-xz}). 

\paragraph{Step 1.}
The first step is to estimate the location of eye in the screen coordinate system.
This can be done with the aid of the user, who is asked to sit at a pre-set distance $z$ from the screen and look perpendicular at the screen plane (such that the angle between the gaze vector and the screen plane becomes 90\textdegree). 
He is then instructed to mark the point of gaze on the screen, denoted by $\{x, y\}$.
In this situation, these marked screen coordinates would directly correspond to the $x$ and $y$ coordinates of the eye in the screen coordinate space.
Thus, the eye location can be determined as: $\mathbf{e}_\texttt{SCS} = \{x^{\texttt{eye}}_{\texttt{SCS}}, y^{\texttt{eye}}_{\texttt{SCS}}, z^{\texttt{eye}}_{\texttt{SCS}}\} = \{x, y, z\}$.

During this time, the rough 3D location of the eye in the camera coordinate system is also obtained (from the eye landmarks of the face modelling step) and represented as $\mathbf{e}_{\texttt{CCS}}$. With this pair of corresponding eye locations obtained, the translation vector $T$ can be expressed by:
\begin{equation}
	\mathbf{e}_\texttt{SCS} =  \mathbf{R} \cdot {\mathbf{e}_{\texttt{CCS}}} + \mathbf{T}
	\quad\implies\quad
	\mathbf{T} = \mathbf{e}_\texttt{SCS} - \mathbf{R} \cdot {\mathbf{e}_{\texttt{CCS}}}.
	\label{eqn:T}
\end{equation}

\paragraph{Step 2.}
Next, without (significantly) changing the head/eye position, the user is asked to look at a different pre-determined point on the screen $\{x_{\texttt{screen}}, y_{\texttt{screen}}\}$.

During this time, the gaze estimation system is used to obtain the gaze direction vector in the camera coordinate system:
\begin{equation}
	\mathbf{g}_{\texttt{CCS}} = \left[\begin{matrix} x^{\texttt{gaze}}_{\texttt{CCS}} && y^{\texttt{gaze}}_{\texttt{CCS}} && z^{\texttt{gaze}}_{\texttt{CCS}} \end{matrix}\right]^T.	
	\label{eqn:g}
\end{equation}
This is a normalized direction vector whose values denote a point on a unit sphere. 
Both the pitch $\alpha$ (about the $x$-axis) and the yaw $\gamma$ (about the $y$-axis) angles of the gaze w.r.t the camera can be re-obtained from this gaze direction vector:
\begin{equation}
	\alpha = \arctan2(-y^{\texttt{gaze}}_{\texttt{CCS}}, z^{\texttt{gaze}}_{\texttt{CCS}})
	\quad\&\quad \gamma = \arctan2(x^{\texttt{gaze}}_{\texttt{CCS}}, z^{\texttt{gaze}}_{\texttt{CCS}}).
	\label{eqn:ag}
\end{equation}

Once $\alpha$ is determined, we can calculate the camera pitch angle $\rho$ between the camera and the screen:
\begin{equation}
	\rho = \arctan(\frac{-y^{\texttt{eye}}_{\texttt{SCS}} + y_{\texttt{screen}}}{z^{\texttt{eye}}_{\texttt{SCS}}}) - \alpha.
\end{equation}
Using this in Equation \ref{eqn:R}, the rotation matrix $\mathbf{R}$ can be fully determined. 
This known $\mathbf{R}$ can now be plugged into Equation \ref{eqn:T} to also determine the translation vector $\mathbf{T}$. 
This procedure can be repeated for multiple calibration points in order to obtain a more robust aggregate estimate of the transformation parameters. 

\paragraph{Step 3.}
Once calibration is complete, any new eye location $\hat{\mathbf{e}}_{\texttt{CCS}}$ can be converted to the screen coordinate space:
\begin{equation}
\hat{\mathbf{e}}_{\texttt{SCS}} = 
	[\hat{x}^{\texttt{eye}}_{\texttt{SCS}}, \hat{y}^{\texttt{eye}}_{\texttt{SCS}}, \hat{z}^{\texttt{eye}}_{\texttt{SCS}}] = \mathbf{R} \cdot \hat{\mathbf{e}}_{\texttt{CCS}} + \mathbf{T}.
\end{equation}
Using the associated new gaze angles $\hat{\alpha}$ and $\hat{\gamma}$, the point of gaze on the screen can be obtained:
\begin{equation}
	\hat{x}_{\texttt{screen}} = \hat{z}^{\texttt{eye}}_{\texttt{SCS}} \cdot tan(\hat{\gamma}) + \hat{x}^{\texttt{eye}}_{\texttt{SCS}}
\quad\&\quad
	\hat{y}_{\texttt{screen}} = \hat{z}^{\texttt{eye}}_{\texttt{SCS}} \cdot tan(\hat{\alpha} + \rho) + \hat{y}^{\texttt{eye}}_{\texttt{SCS}},
	\label{eqn:screen}
\end{equation}

\subsection{Machine Learning~(ML)-based Calibration.}
Since the task of gaze vector to gaze point calibration requires learning the mapping between two sets of coordinates, this can be treated as a regression problem.
In our implementation, we use a linear ridge regression model for this task for it's ability to avoid overfitting when training samples are scarce.
The input to this calibration model includes the predicted gaze-vector angles and the 3D location of the eye, all in the camera coordinate system.
The outputs are the 2D coordinates of the gaze-point on the screen in the screen coordinate system.

During calibration, the user is asked to look at a number of predefined points on the screen (such that they span the full region of the screen) while their gaze and eye locations are estimated and recorded for each of these points.
These calibration samples are then used to train the model.
Given enough training/calibration points, this model is expected to implicitly learn the mapping between the two coordinate systems. 

\subsection{`Hybrid' Geometric Regression Calibration.}
To combine the benefits of geometry based prior knowledge with ML based regression, a hybrid geometric regression technique can be derived where machine learning is used to infer the required geometric transformation parameters.

As before, we assume the roll and yaw angles between the camera and the screen are 0\textdegree. 
The only unknown between the pose of the camera w.r.t the screen is the pitch angle $\rho$.
The rotation and translation matrices are the same as given by Equation \ref{eqn:R}, and the formulations of gaze pitch and yaw angles $\alpha$ and $\gamma$ stay the same as defined by Equation \ref{eqn:ag}.


Again, during calibration the user is asked to look at a number of varied predefined points on the screen while their gaze directions and eye locations are recorded. 
These data samples are then used to jointly minimize the reprojection errors (squared Euclidean distance) to learn the required transformation parameters $\rho, \Delta x , \Delta y, \Delta z$:
\begin{equation}
	\argmin_{\rho, \Delta x, \Delta y, \Delta z} \sum_{i=1}^{N} 
	\left((x_{\texttt{point}}^i - \hat{x}_{\texttt{screen}}^i)^2 
		+ (y_{\texttt{point}}^i - \hat{y}_{\texttt{screen}}^i)^2 \right),
\end{equation}
where $N$ is the number of training/calibration points; $\{x_{\texttt{point}}, y_{\texttt{point}}\}$ denote the ground truth screen points, while $\{\hat{x}_{\texttt{screen}}, \hat{y}_{\texttt{screen}}\}$ are the predicted gaze points on screen as estimated using Equation \ref{eqn:screen}.
\\We solve this minimization problem by differential evolution~\cite{Storn1997}.

	\section{Experiments and Results}
\label{sec:exp}

\subsection{Datasets}
We perform all our experiments on two publicly available gaze-tracking datasets:

\paragraph{MPIIFaceGaze~\cite{written-on-your-face}.}
This dataset is an extended version of MPIIGaze~\cite{MPII} with available human face region.
It contains 37,667 images from 15 different participants.
The images have variations in illumination, personal appearance, head pose and camera-screen settings.
The ground truth gaze target on the screen is given as a 3D point in the camera coordinate system.

\paragraph{EYEDIAP~\cite{eyediap}.}
This dataset provides 94 video clips recorded in different environments from 16 participants.
It has two kinds of gaze targets: screen points and 3D floating targets.
It also has two types of head movement conditions: static and moving head poses.
For our experiments, we choose the screen point target with both the static and moving head poses, which contains 218,812 images.

\subsection{Exp 1: Speed/Accuracy Trade-off for Varying Input Sizes}

\subsubsection*{Setup.}
\begin{figure}
	\floatbox[{\capbeside\thisfloatsetup{capbesideposition={right,bottom},capbesidewidth=0.5\textwidth}}]{figure}[\FBwidth]
	{\caption{Examples of three input types used in the experiments: (left) face crop, sized 224$\times$224 and 112$\times$112; (right top) two eyes region crop, sized 180$\times$60 and 90$\times$30; (right bottom) single eye crop, sized 60$\times$32 and 30$\times$18.}
	\label{fig:examples}}
	{
		\centering
		\includegraphics[width=0.4\textwidth]{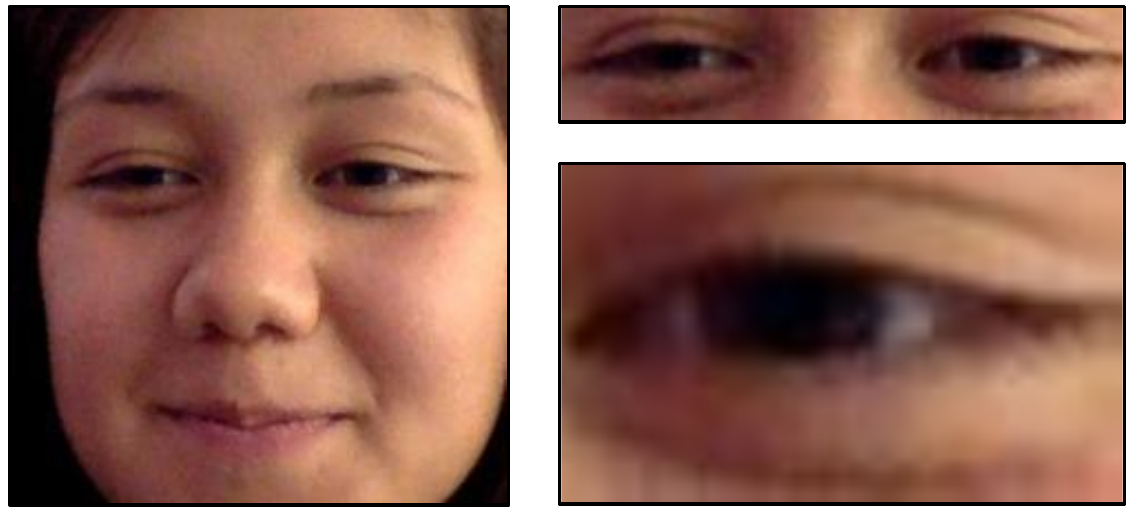}
	}
\end{figure}

A gaze vector can be predicted based on various input image sizes, as illustrated in Figure~\ref{fig:examples}:
\begin{itemize}
	\item Full face image: The largest and most informative input;
	\item Two eyes: Medium sized and informative;
	\item Single eye: Minimal information and smallest size.
\end{itemize}
To assess the performance gains of different input sizes vs their computational loads and accuracy, we setup an experiment where we vary the input training and testing data to the neural network while keeping all other settings fixed.
We then measure the accuracy of the system and compute their individual inference-time computational loads.

For this experiment, we individually train our deep network on each of the multiple types and sizes of the pre-processed inputs shown in Figure~\ref{fig:examples}.
In order to obtain a reliable error metric, we perform 5-fold cross-validation training.
This experiment is repeated for both the MPIIFaceGaze and EYEDIAP dataset.

\subsubsection*{Results.}
\begin{figure}
	\centering
	\subfloat[][MPIIFaceGaze]{\includegraphics[width=.5\textwidth]{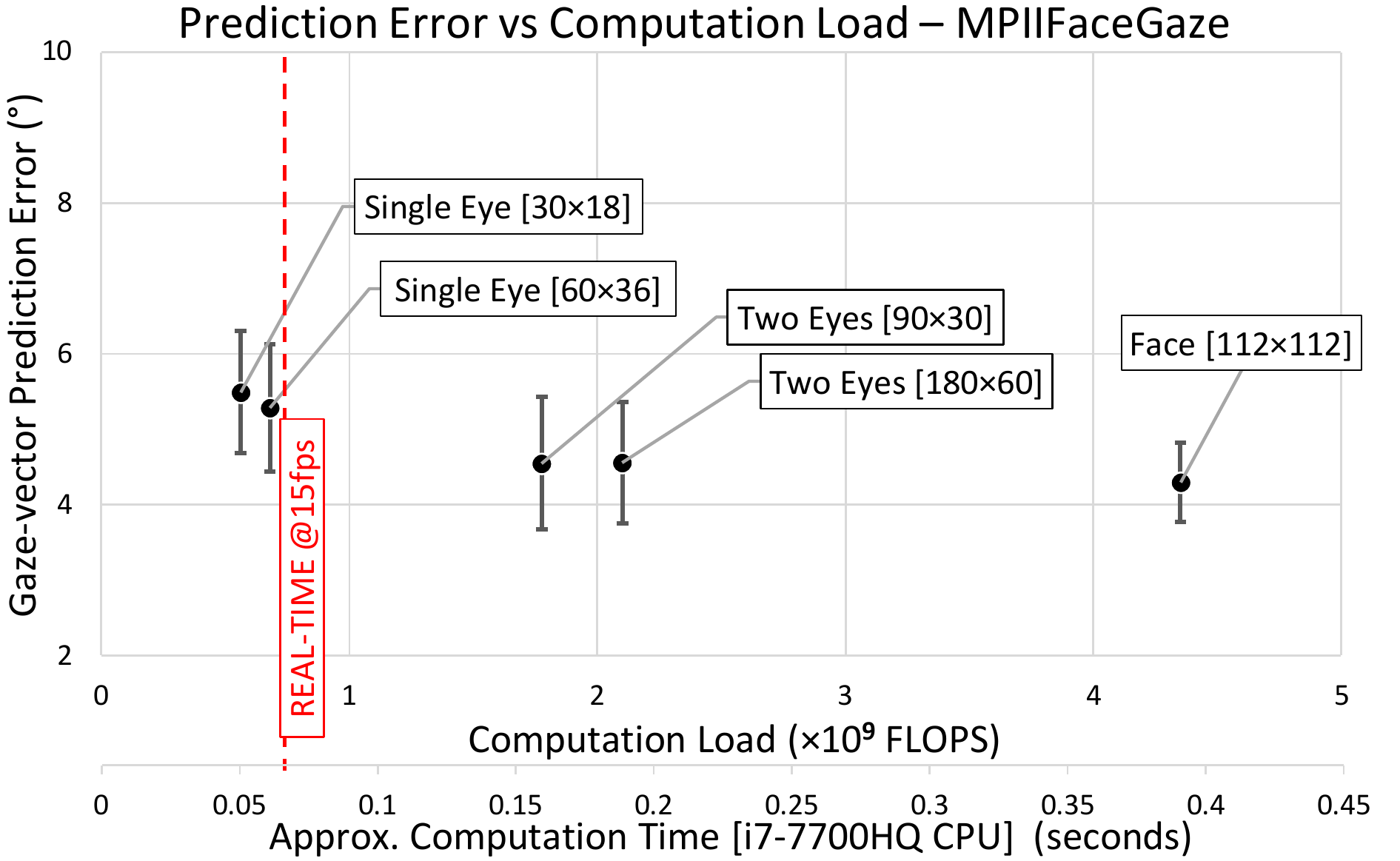}\label{fig:inputtype-mpii}}
	\subfloat[][EYEDIAP]{\includegraphics[width=.5\textwidth]{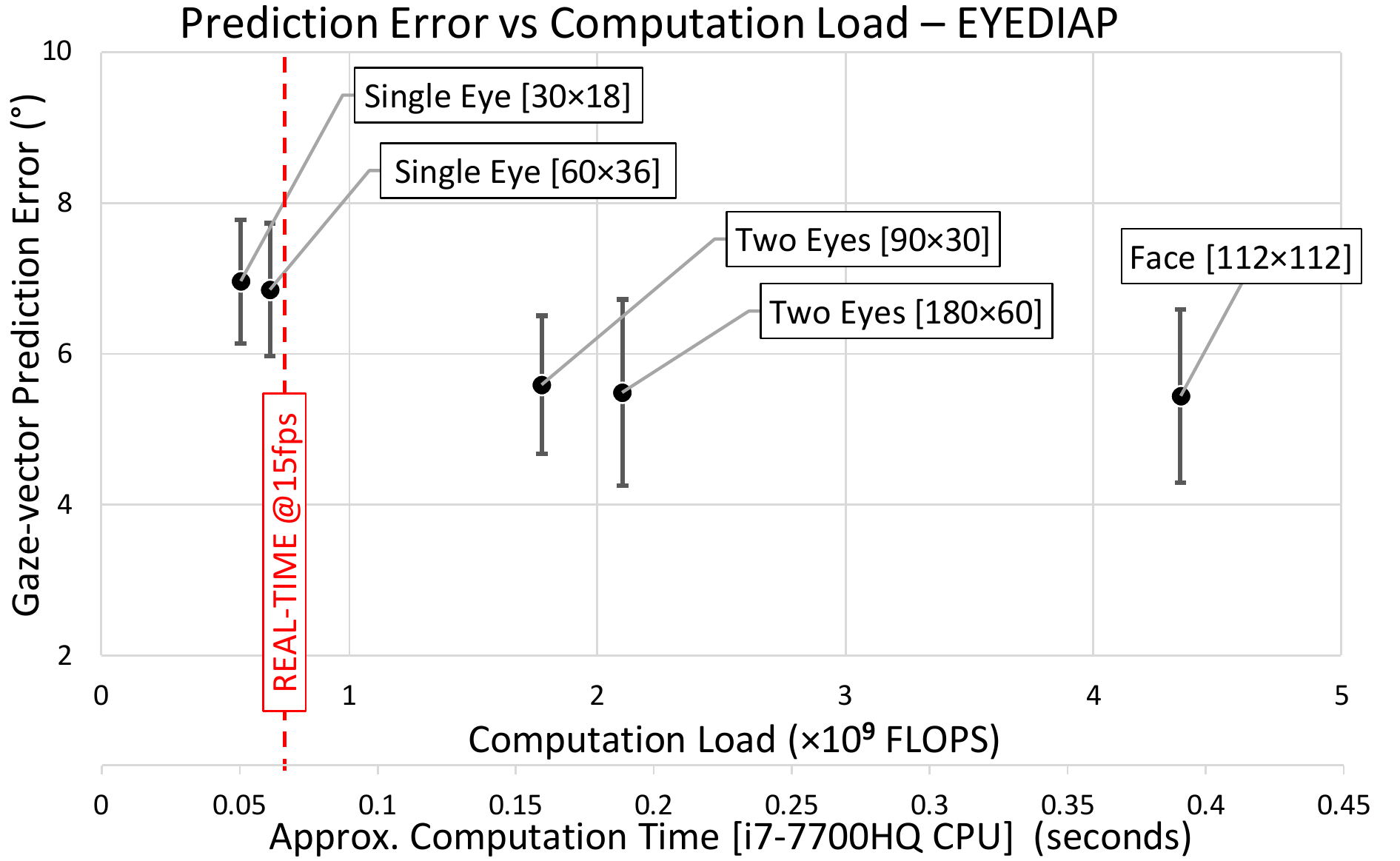}\label{fig:inputtype-eyediap}}
	{\caption{Scatter plots of the performance of a VGG16 based gaze tracking network trained on different input types vs their computation load/time in FLOPS/seconds.
	The error bars represent the standard deviation of the errors (5-fold cross-validation).
	While the computation cost of these inputs vastly vary, they all perform in roughly the same range of accuracy.
	The red dashed line represents approximate real-time computation at 15~fps on an \emph{Intel i7} CPU.}
	\label{fig:inputtype}}
\end{figure}

The results of this experiment can be seen in Figure~\ref{fig:inputtype}.
As expected, we observed that the lowest error rates are obtained by the largest size of input data with the maximum amount of context: the full face image.
We also observe that using this input type results in the highest amount of computation load.

As we reduce the input sizes, the accuracy only marginally degrades while the computation load gets cut down severely. 
In fact, even if we simply use a crop of the eye region or just the crop of a single eye, we obtain accuracies comparable to that from full face input albeit with a fraction of the computation.

\subsection{Exp 2: Usability/Accuracy Trade-off for Screen Calibration}

\subsubsection*{Setup.}
To evaluate the three screen calibration techniques proposed, we train and test them individually using calibration data samples from MPIIFaceGaze and EYEDIAP.
This data for screen calibration consists of pairs of gaze-vectors and their corresponding ground truth screen points.
Using this, calibration methods are trained to predict the 2D screen points from the 3D gaze-vectors. 
We evaluate on noise-free ground truth gaze-vectors and on realistic predicted gaze-vectors (using 30$\times$18 eye crop as input) so as to assess the accuracy of the complete camera-to-screen eye-tracking pipeline. 
As training data, we obtain calibration data pairs of gaze-vectors and points such that they are spread out evenly over the screen area.
This is done by dividing the screen in an evenly-spaced grid and extracting the same number of points from each grid region.

\subsubsection*{Results.}
The results of these experiments can be seen in Table~\ref{tab:calresults} for a fixed calibration training set size of 100 samples.
For the `theoretical' task of predicting gaze-points from noise-free ground truth gaze-vectors, we see in Table~\ref{tab:calresults-gt} that the hybrid geometric regression method outperforms others.
We see that the gap in performance is smaller when head poses are static, while the hybrid method does better for moving head poses.
This suggests that for the simplest evaluation on static head poses with noise-free gaze-vectors, all methods perform well; however, as movement is introduced, the limitations of the purely geometric method and the advantage of hybrid method becomes clearer.
\begin{table}
	\centering
	\subfloat[][Groundtruth gaze-vector calibration]{\includegraphics[width=.46\textwidth, trim=0.6cm 0 0 0, clip]{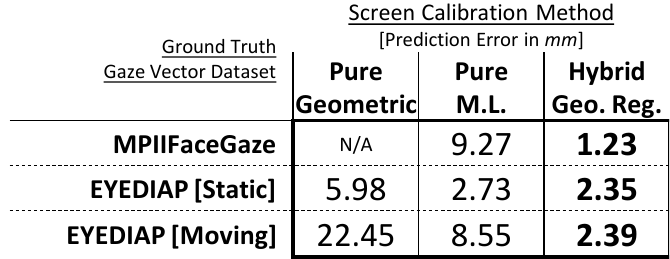}\label{tab:calresults-gt}} \quad\quad
	\subfloat[][Predicted gaze-vector calibration]{\includegraphics[width=.46\textwidth, trim=0.6cm 0 0 0, clip]{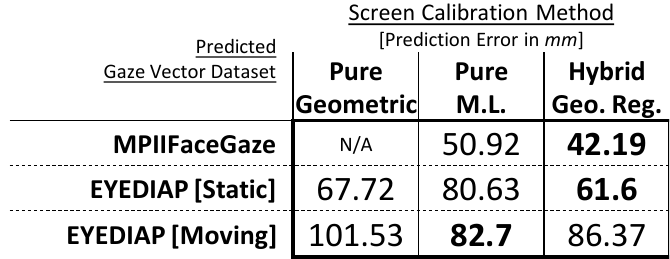}\label{tab:calresults-pred}}
	{\caption{Performance of calibration methods (trained with 100 samples) on different datasets and conditions expressed in gaze-point prediction errors (in \textit{mm}). Hybrid geometric regression technique significantly outperforms both purely geometric and purely machine learning (M.L.) based calibration methods in most conditions.
	\newline Legend: [Static] denotes static head poses, [Moving] denotes moving head poses.}
	\label{tab:calresults}}
\end{table}

When calibration is performed on actual gaze-vectors predicted by the system, overall accuracy deteriorates by one to three orders of magnitude. Comparing the methods, the hybrid geometric regression method also does well compared to others in most conditions, as seen in Table~\ref{tab:calresults-pred}.
We observe that the purely geometric method actually copes better than the ML based method when head poses are static. 
However, it's performance severely degrades with moving head poses. 
Also, the ML method is able to marginally outperform the hybrid method on moving head poses. This is likely because given sufficient training samples, the ML method is able to learn features from the input that the other---more rigid---methods cannot do.
Note that only the EYEDIAP dataset results are reported for the purely geometric technique, since this method can only be trained on static head poses and MPIIFaceGaze does not have any static head poses (the geometric method can still be tested on the moving head poses of EYEDIAP). 

\begin{figure}
	\centering
	\subfloat[][MPIIFaceGaze]{\includegraphics[width=.45\textwidth]{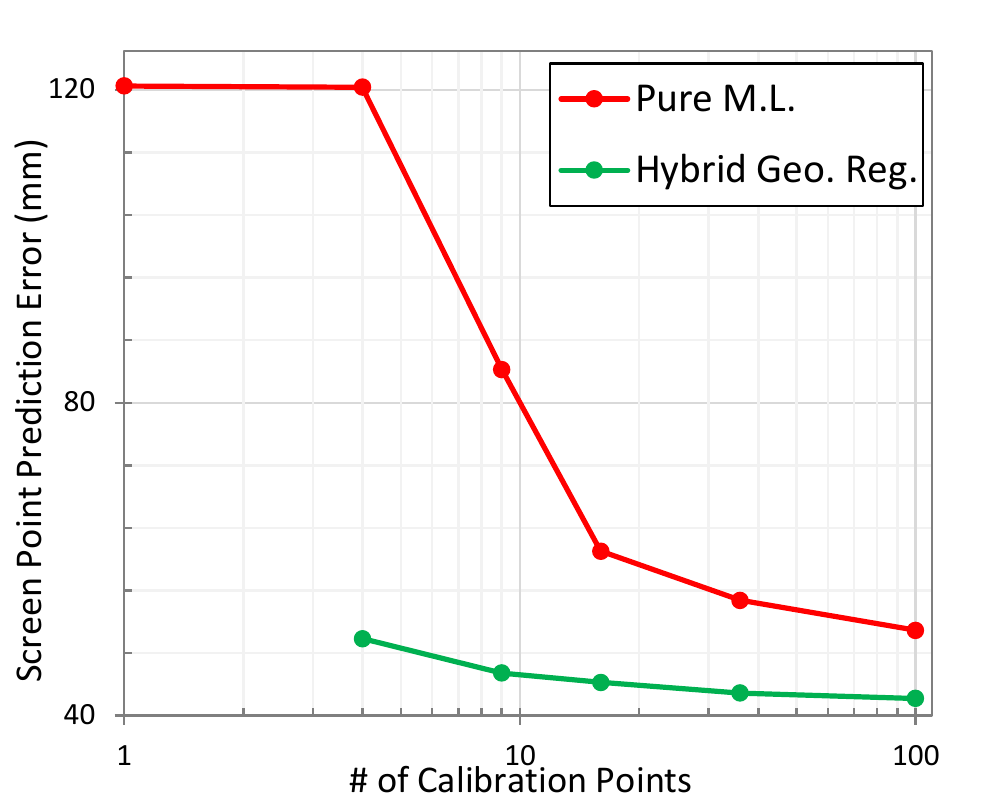}\label{fig:calresults-mpii}}\qquad
	\subfloat[][EYEDIAP (Moving)]{\includegraphics[width=.45\textwidth]{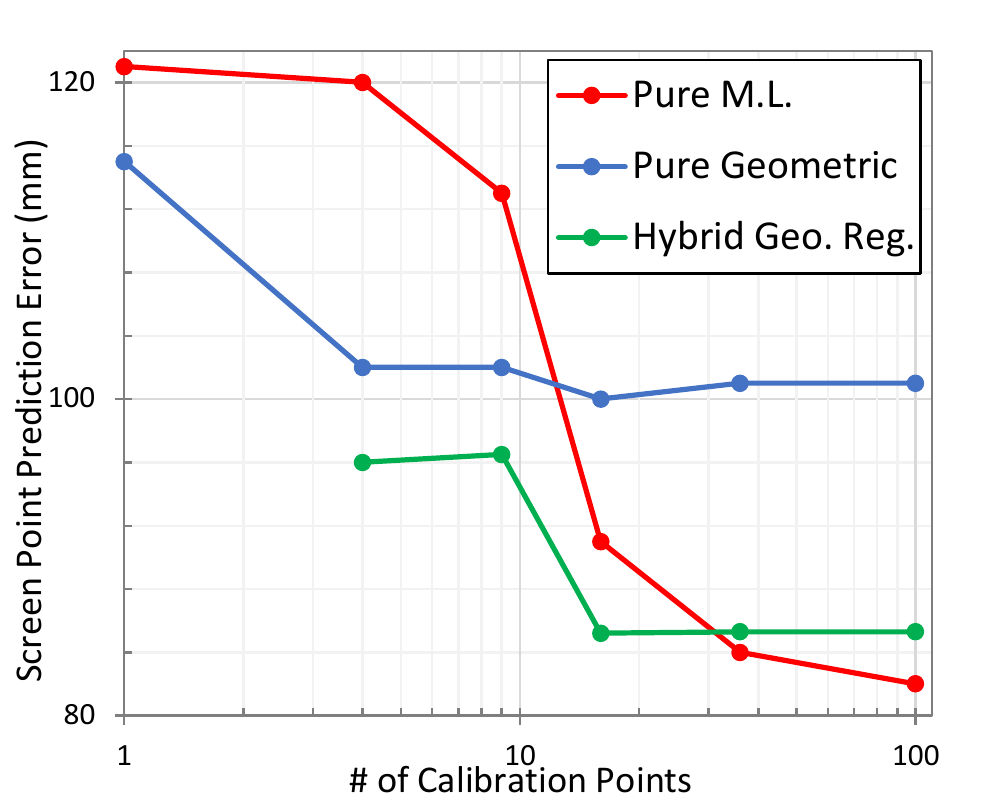}\label{fig:calresults-eyediap}}
	\caption{Learning curves of the calibration techniques on MPIIFaceGaze and EYEDIAP dataset ($\log$ scale). 
	The purely geometric method performs better than ML method when calibration data is scarce, but does not improve further when more data is available.
	The ML method improves greatly when calibration data becomes abundant. 
	The hybrid geometric regression method performs the best over a wide range of calibration data points used.}
	\label{fig:calresults-indepth}
\end{figure}

To assess the efficiency of these calibration methods, we must ascertain the least amount of calibration samples required with which satisfactory performance can still be attained.
This can be assessed by observing the learning curves of the calibration methods, where the prediction errors of the methods are plotted against the number of calibration/training points used.
This is shown in Figure~\ref{fig:calresults-indepth}.

The hybrid method is able to outperform both the other methods even when a  low number of calibration points are available.
An interesting observation seen in Figure~\ref{fig:calresults-eyediap} is that the purely geometric method actually performs  better than the ML method when the number of calibration points is  low ($\lessapprox$ 9).
This can be due to the rigid and pre-defined nature of the geometric model which has prior knowledge strongly imparted into it. 
On the other hand, the ML model requires more data points to learn the underlying geometry from scratch.
This is also seen in the results: as the number of points increase, the ML model's performance improves while the geometric model stagnates.
Overall, the lower error rate of the hybrid model over a broad range of used calibration points affirms its strengths over the overtly rigid geometric model and the purely data-driven ML approach.

	\section{Discussion}
\label{sec:disc}
The experiments related to input types and sizes produce some insightful and promising results. 
The comparison between them with respect to their performance vs their computational load indicate that the heavier processing of larger inputs with more contextual information is not worth the performance gain they produce.
Roughly the same accuracies can be obtained by a system that relies only on eye image crops. 
In contrast, the gap in the computational load between these two input types is a factor of 20.
This supports our idea that for an objective measurement task like gaze-vector prediction, the value of context is limited.
These results can help in guiding the design of eye tracking systems meant for real-time applications where efficiency is key.

\bigskip
Outputs in the form of gaze-vectors are not always readily useful in a computer-facing scenario: they need to be projected onto the screen to actually determine where the person is looking.
This area has received little attention in literature, and our experiments provide some insight. 
Our comparison of three calibration techniques show that a hybrid geometric regression method gives the overall best performance over a wide range of available calibration data points.
Our results show that purely geometric modelling works better when calibration points are very few, while a purely ML method outperforms it when more points become available.
However, a hybrid model offers a robust trade-off between them. 

\section{Conclusion}
\label{sec:conc}
In this work, we explored the value of visual context in input for the task of gaze tracking from camera images.
Our study gives an overview of the accuracy different types and sizes of inputs can achieve, in relation to the amount of computation their analysis requires.
The results strongly showed that the improvement obtained from large input sizes with rich contextual information is limited while their computational load is prohibitively high. 
Additionally, we explored three screen calibration techniques that project gaze-vectors onto screens without knowing the exact transformations, achieved with the cooperation of the user.
We showed that in most cases, a hybrid geometric regression method outperforms a purely geometric or machine learning based calibration while generally requiring less calibration data points and thus being more efficient.

	\section*{Acknowledgement}
	The authors are grateful to Messrs. Tom Viering, Nikolaas Steenbergen, Mihail Bazhba, Tim Rietveld, Hans Tangelder for their most valuable inputs.
	\\Special thanks to Sig. Fabio Gatti for critical driving services :)

	\bibliographystyle{splncs04} 
	\bibliography{../../references}

\begin{thebibliography}{10}
\providecommand{\url}[1]{\texttt{#1}}
\providecommand{\urlprefix}{URL }
\providecommand{\doi}[1]{https://doi.org/#1}

\bibitem{CCNFs}
Baltru{\v{s}}aitis, T., Robinson, P., Morency, L.P.: Continuous conditional
  neural fields for structured regression. In: Proceedings of the European
  Conference on Computer Vision. pp. 593--608 (2014)

\bibitem{dilated_conv_gaze}
Chen, Z., Shi, B.E.: Appearance-based gaze estimation using
  dilated-convolutions. In: Proceedings of the Asian Conference on Computer
  Vision. pp. 309--324 (2018)

\bibitem{asymmetric}
Cheng, Y., Lu, F., Zhang, X.: Appearance-based gaze estimation via
  evaluation-guided asymmetric regression. In: Proceedings of The European
  Conference on Computer Vision (2018)

\bibitem{Monocular}
Deng, H., Zhu, W.: Monocular free-head 3d gaze tracking with deep learning and
  geometry constraints. In: Proceedings of the International Conference on
  Computer Vision. pp. 3162--3171 (2017)

\bibitem{RT-GENE}
Fischer, T., Jin~Chang, H., Demiris, Y.: Rt-gene: Real-time eye gaze estimation
  in natural environments. In: Proceedings of the European Conference on
  Computer Vision (2018)

\bibitem{eyediap}
Funes~Mora, K.A., Monay, F., Odobez, J.M.: Eyediap: A database for the
  development and evaluation of gaze estimation algorithms from rgb and rgb-d
  cameras. In: Proceedings of the ACM Symposium on Eye Tracking Research and
  Applications (2014)

\bibitem{DenseNet}
Huang, G., Liu, Z., Van Der~Maaten, L., Weinberger, K.Q.: Densely connected
  convolutional networks. In: Proceedings of the IEEE conference on Computer
  Vision and Pattern Recognition. pp. 4700--4708 (2017)

\bibitem{DBLP:journals/corr/IoffeS15}
Ioffe, S., Szegedy, C.: Batch normalization: Accelerating deep network training
  by reducing internal covariate shift. In: International Conference on Machine
  Learning. pp. 448--456 (2015)

\bibitem{Kasprowski2014}
Kasprowski, P., Harezlak, K., Stasch, M.: Guidelines for the eye tracker
  calibration using points of regard. In: Information Technologies in
  Biomedicine, Volume 4. pp. 225--236. Springer International Publishing (2014)

\bibitem{kingma2014adam}
Kingma, D.P., Ba, J.: Adam: A method for stochastic optimization. In:
  Proceedings of the International Conference on Learning Representations
  (2015)

\bibitem{gazecapture}
Krafka, K., Khosla, A., Kellnhofer, P., Kannan, H., Bhandarkar, S., Matusik,
  W., Torralba, A.: Eye tracking for everyone. In: IEEE Conference on Computer
  Vision and Pattern Recognition (CVPR) (2016)

\bibitem{hourglass}
Newell, A., Yang, K., Deng, J.: Stacked hourglass networks for human pose
  estimation. In: European conference on computer vision. pp. 483--499.
  Springer (2016)

\bibitem{Recurrent_Gaze}
Palmero, C., Selva, J., Bagheri, M.A., Escalera, S.: Recurrent {CNN} for 3d
  gaze estimation using appearance and shape cues. In: British Machine Vision
  Conference (BMVC) (2018)

\bibitem{webgazer}
Papoutsaki, A., Sangkloy, P., Laskey, J., Daskalova, N., Huang, J., Hays, J.:
  Webgazer: Scalable webcam eye tracking using user interactions. In:
  Proceedings of the International Joint Conference on Artificial Intelligence.
  pp. 3839--3845 (2016)

\bibitem{park2019few}
Park, S., Mello, S.D., Molchanov, P., Iqbal, U., Hilliges, O., Kautz, J.:
  Few-shot adaptive gaze estimation. In: Proceedings of the IEEE International
  Conference on Computer Vision. pp. 9368--9377 (2019)

\bibitem{Pictorial}
Park, S., Spurr, A., Hilliges, O.: Deep pictorial gaze estimation. In: European
  conference on computer vision (2018)

\bibitem{Head_Pose_Invariant_Gaze_Tracking}
Ranjan, R., De~Mello, S., Kautz, J.: Light-weight head pose invariant gaze
  tracking. In: Proceedings of the IEEE Conference on Computer Vision and
  Pattern Recognition Workshops. pp. 2156--2164 (2018)

\bibitem{Rodrigues2010CameraPE}
Rodrigues, R., Barreto, J.P., Nunes, U.: Camera pose estimation using images of
  planar mirror reflections. In: European Conference on Computer Vision (2010)

\bibitem{ILSVRC15}
Russakovsky, O., Deng, J., Su, H., Krause, J., Satheesh, S., Ma, S., Huang, Z.,
  Karpathy, A., Khosla, A., Bernstein, M., Berg, A.C., Fei-Fei, L.: {ImageNet
  Large Scale Visual Recognition Challenge}. International Journal of Computer
  Vision (IJCV) pp. 211--252 (2015)

\bibitem{vgg}
Simonyan, K., Zisserman, A.: Very deep convolutional networks for large-scale
  image recognition. In: International Conference on Learning Representations
  (2015)

\bibitem{Storn1997}
Storn, R., Price, K.: Differential evolution -- a simple and efficient
  heuristic for global optimization over continuous spaces. Journal of Global
  Optimization pp. 341--359 (1997)

\bibitem{Image_Normalized}
{Sugano}, Y., {Matsushita}, Y., {Sato}, Y.: Learning-by-synthesis for
  appearance-based 3d gaze estimation. In: Proceedings of the IEEE Conference
  on Computer Vision and Pattern Recognition. pp. 1821--1828 (2014)

\bibitem{tan2002appearance}
Tan, K.H., Kriegman, D.J., Ahuja, N.: Appearance-based eye gaze estimation. In:
  Sixth IEEE Workshop on Applications of Computer Vision, 2002.(WACV 2002).
  Proceedings. pp. 191--195. IEEE (2002)

\bibitem{GPR}
Tripathi, S., Guenter, B.: A statistical approach to continuous
  self-calibrating eye gaze tracking for head-mounted virtual reality systems.
  In: 2017 IEEE Winter Conference on Applications of Computer Vision (WACV).
  pp. 862--870. IEEE (2017)

\bibitem{3dmm}
Wood, E., Baltru\v{s}aitis, T., Morency, L.P., Robinson, P., Bulling, A.: A 3d
  morphable model of the eye region. In: Proceedings of the 37th Annual
  Conference of the European Association for Computer Graphics: Posters. pp.
  35--36 (2016)

\bibitem{EyeTab}
Wood, E., Bulling, A.: Eyetab: Model-based gaze estimation on unmodified tablet
  computers. In: Proceedings of the Symposium on Eye Tracking Research and
  Applications. pp. 207--210. ACM (2014)

\bibitem{yu2019improving}
Yu, Y., Liu, G., Odobez, J.M.: Improving few-shot user-specific gaze adaptation
  via gaze redirection synthesis. In: Proceedings of the IEEE Conference on
  Computer Vision and Pattern Recognition. pp. 11937--11946 (2019)

\bibitem{written-on-your-face}
Zhang, X., Sugano, Y., Fritz, M., Bulling, A.: It’s written all over your
  face: Full-face appearance-based gaze estimation. In: IEEE Conference on
  Computer Vision and Pattern Recognition Workshops (CVPRW). pp. 2299--2308
  (2017)

\bibitem{in_the_wild}
Zhang, X., Sugano, Y., Fritz, M., Bulling, A.: Appearance-based gaze estimation
  in the wild. In: Proceedings of the IEEE conference on computer vision and
  pattern recognition. pp. 4511--4520 (2015)

\bibitem{MPII}
Zhang, X., Sugano, Y., Fritz, M., Bulling, A.: Mpiigaze: Real-world dataset and
  deep appearance-based gaze estimation. IEEE Transactions on Pattern Analysis
  and Machine Intelligence  \textbf{41}(1),  162--175 (2017)

\end{thebibliography}
\end{document}